\documentclass[conference]{IEEEtran}
\IEEEoverridecommandlockouts
\usepackage{cite}
\usepackage{amsmath,amssymb,amsfonts}
\usepackage{algorithmic}
\usepackage{graphicx}
\usepackage{textcomp}
\usepackage{xcolor}
\usepackage{tabularx}
\usepackage{multirow}
\usepackage{multicol}
\usepackage{array}
\usepackage{bm}
\usepackage[misc]{ifsym}
\usepackage{subfigure}
\usepackage{threeparttable}

\def\BibTeX{{\rm B\kern-.05em{\sc i\kern-.025em b}\kern-.08em
    T\kern-.1667em\lower.7ex\hbox{E}\kern-.125emX}}

\begin{document}

\title{A Novel Multi-scale Dilated 3D CNN for Epileptic Seizure Prediction}

\author{\IEEEauthorblockN{Ziyu Wang, Jie Yang and Mohamad Sawan}
	\IEEEauthorblockA{ Cutting-Edge Net of Biomedical Research and INnovation (CenBRAIN), \\
		School of Engineering, Westlake University, Hangzhou, China \\
		{yangjie@westlake.edu.cn}
	}
}

\maketitle

\begin{abstract}
Accurate prediction of epileptic seizures allows patients to take preventive measures in advance to avoid possible injuries. In this work, a novel convolutional neural network (CNN) is proposed to analyze time, frequency, and channel information of electroencephalography (EEG) signals. The model uses three-dimensional (3D) kernels to facilitate the feature extraction over the three dimensions. The application of multi-scale dilated convolution enables the 3D kernel to have more flexible receptive fields. The proposed CNN model is evaluated with the CHB-MIT EEG database, the experimental results indicate that our model outperforms the existing state-of-the-art, achieves 80.5\% accuracy, 85.8\% sensitivity and 75.1\% specificity.
\end{abstract}

\section{Introduction}
\label{sec:intro}

Epilepsy is a neurological disorder disease, characterized by unexpected recurrent attacks, affecting 1\% population in the world \cite{RN1} \cite{RN2}. Due to the uncertainty of seizure occuriance, it has a great impact on patients’ daily life, and can even threaten a patients' life if they are in some dangerous situation (e.g. driving, going up or down stairs) when a seizure occurs. Thus, accurate seizure prediction is essential to help the patients avoid possible injuries. EEG is the most commonly used data to study epileptic seizures, which could be categorized into four types according to different states of represented human brain activity, namely preictal (before seizure), ictal (seizure), postictal (after seizure) and interictal (normal stage) \cite{RN3} \cite{RN4} \cite{RN5} \cite{RN6}. By analyzing EEG signals with machine learning or deep learning algorithms, a distinction between preictal and interictal can be achieved, thus enables seizure prediction \cite{RN7} \cite{RN8} \cite{8572221}. 

Machine learning algorithms rely on hand-craft features, raw EEG signals need to be preprocessed by multiple filtering techniques to remove the noise and artifacts. Bandpass filter, adaptive filter, Kalman filter, Wiener filter and Bayes filter are most commonly used filtering techniques \cite{RN10} \cite{9173735}. Univariate spectral power \cite{RN16}, spike rate \cite{RN21}, power spectral \cite{RN24}, permutation entropy \cite{RN9} and bi-spectral entropy \cite{RN170}, for example, are the features that have been used in previous machine learning-based works. These features are further fed into a classifier (e.g. Support vector machine (SVM), Multilayer perceptron (MLP), Decision trees) to get the prediction result. Although
\begin{figure}[thb]
	\setlength{\belowcaptionskip}{-100pt}
	\begin{center}
		\includegraphics[width=0.48\textwidth]{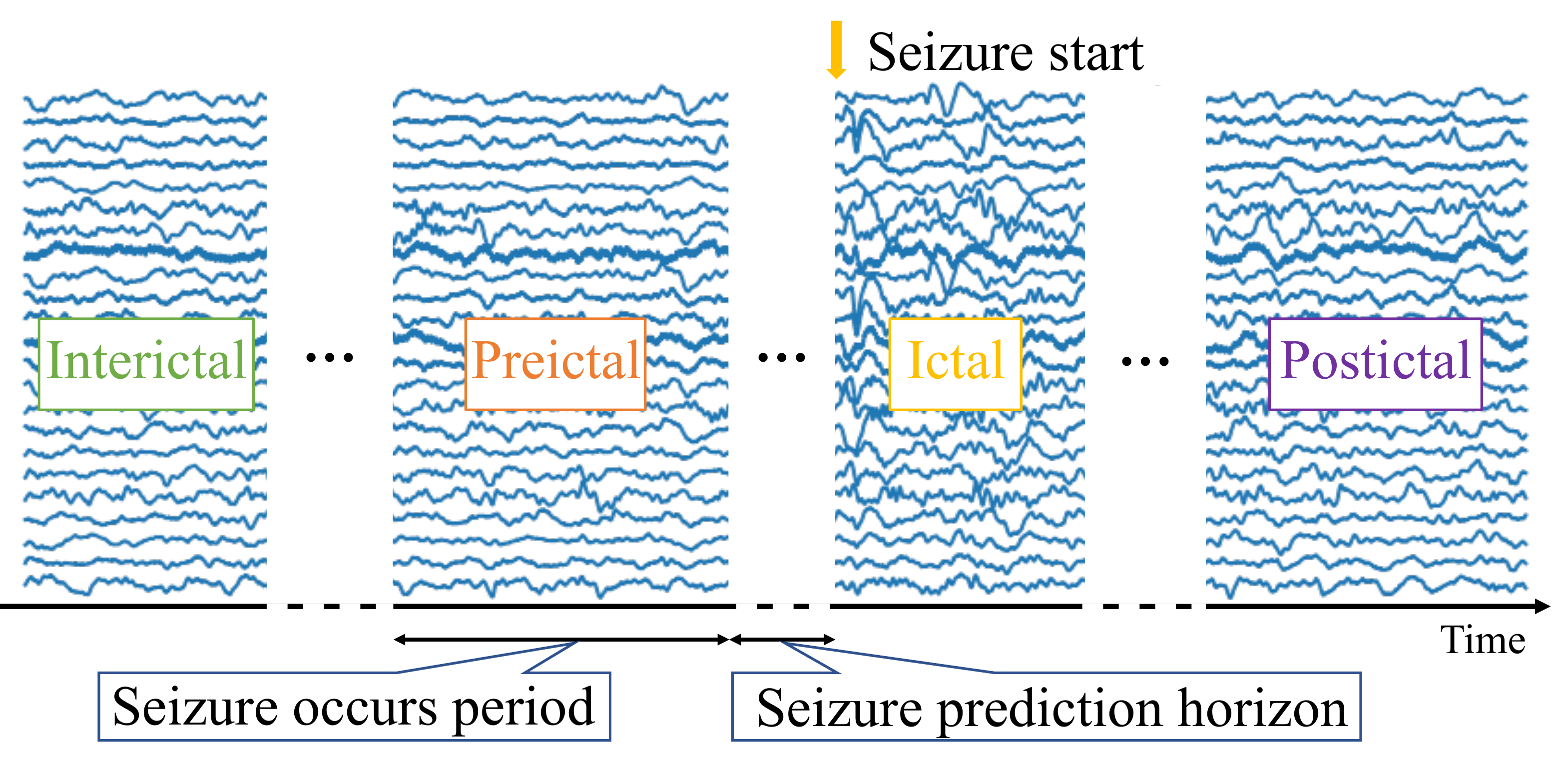}
		\caption{Relationship of interictal, preictal, ictal and postictal. Ictal is during seizures, preictal is a time period precedes ictal, which is before seizure prediction horizon and within seizure occurs period. Postictal refers to the state after seizures, and interictal corresponds to normal stages.}
		\label{relationship}
	\end{center}
\vspace{-0.5cm}
\end{figure}
machine learning methods are feasible, manual processing of EEG signals and complex feature engineering make it difficult to apply in real-time. On the contrary, deep learning is based on CNN, which could extract features automatically on raw data or minimally processed data \cite{RN27}. CNN consists of both modules of feature extraction and classification, which enables an end-to-end training process. However, in many deep learning-based works, the sizes of convolution kernels are small, which only focus on local spatial features. When applied to time series data such as EEG signals, the small receptive field results in the loss of long-term temporal information, hence, affects the prediction performance significantly. 

This study proposes a novel seizure prediction algorithm, it first convert and stack EEG signal into a 3D tensor by short time Fourier transform (STFT) to include time, frequency and channel information. Then 3D convolution kernel \cite{6165309} is utilized to instead of 2D one, as it could extract features from three dimensions simultaneously. To further improve the performance, dilated convolution \cite{2015arXiv151107122Y} is introduced to increase the receptive field. By combining the output from different sizes of dilated 3D convolution kernels, information from multiple time scales could be obtained, which is more suitable for processing long-term time series data.

The remaining sections of this paper are organized as follows: Section \ref{sec:data} introduces the methods of data segmentation and sampling, presents the structure of proposed CNN model. Section \ref{sec:result} gives the evaluation result of proposed model and the comparison with existing models. Section \ref{sec:conclusion} concludes all the contributions in this work.

\section{Materials and methods}
\label{sec:data}

\subsection{Data segmentation and sampling}
\label{sec:data segmentation}
\begin{figure}[t]
	\begin{center}
		\includegraphics[width=0.48\textwidth]{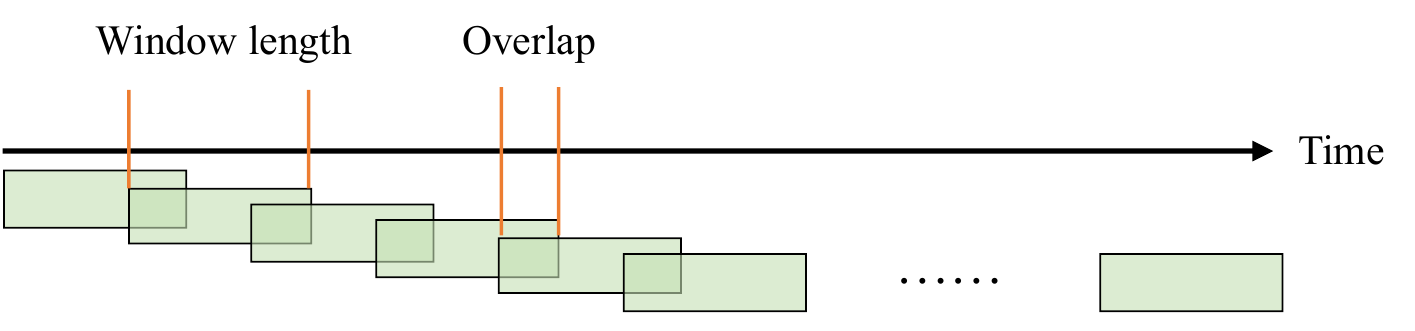}
		\caption{Moving window sampling method.}
		\label{moving window}
	\end{center}
\end{figure}

In order to distinguish between preictal and interictal, the location of these two segments need to be determined on the long-term EEG signal first. As shown in Fig. \ref{relationship}, preictal is a period before ictal (seizure), which depends on seizure start time, seizure occurs period (SOP) and seizure prediction horizon (SPH). According to Maiwald et al., the definition of SOP and SPH is: after a alarm signal, during SPH, no seizure has occurred yet; during SOP, a seizure occurs \cite{RN11}. Based on previous works, SOP and SPH are set as 30 min and 5 min respectively, interictal is defined as 4 h before or after a seizure \cite{RN19} \cite{RN12}. Each preictal or interictal segment is further divided into smaller samples by moving window sampling method, the window length is 30 s, with 8 s overlap to augment the sample set (Fig. \ref{moving window}). Then STFT method is utilized to convert the non-Gaussian and non-stationary EEG signals into time-frequency domain to observe the sudden changes in frequency. The converted sample has a 3D shape, which is (channels, frequency, time).

\begin{figure}[t]
	\begin{center}
		\subfigure[]{
			\includegraphics[width=1.9cm]{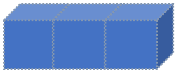}
			\label{standard kernel}
		}
		\subfigure[]{
			\includegraphics[width=6.1cm]{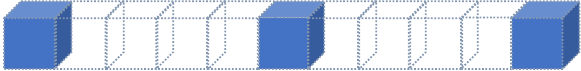}
			\label{dilated kernel}
		}
		\caption{(a) Standard convolution kernel and (b) Dilated convolution kernel, dilated size is 5.}
		\label{kernels}
	\end{center}
\vspace{-0.4cm}
\end{figure}

\subsection{Model structure}
\label{sec:model}
CNN generally refers to 2D CNN and is usually used to process images, but as mentioned in Section \ref{sec:data segmentation}, after STFT the EEG sample has three dimensions: channels, frequency and time. Thus, 3D convolution is utilized to design our network. It is commonly used in video analysis and action recognition in computer vision and is good at solving time series problems. Convolutional neural networks focus on local features, as it is originally applied to images, where the neighboring pixels have a strong correlation. However, raw EEG signal is anisotropy, the corresponding electrodes of two neighboring channels in EEG signals may be far apart on the patient’s scalp. Thus, we use dilated convolution kernel instead of the standard one to analyze EEG signals. Taking the 1 $\times $ 3 kernel as an example, Fig. \ref{standard kernel} is the standard convolution kernel, Fig. \ref{dilated kernel} is the dilated convolution kernel. It can be noted that, there are some spaces between each unit of the dilated kernel (number of spaces=dilated size-1). The dilated convolution kernel can enlarge receptive field without increasing computation as those spaces have no weights and do not participate in convolution calculations. By combining the features extracted by 3D convolution kernels of different dilated sizes, not only the information from neighboring channels and time steps could be obtained, but also those from distant ones.

\begin{figure}[t]
	\begin{center}
		\includegraphics[width=0.48\textwidth]{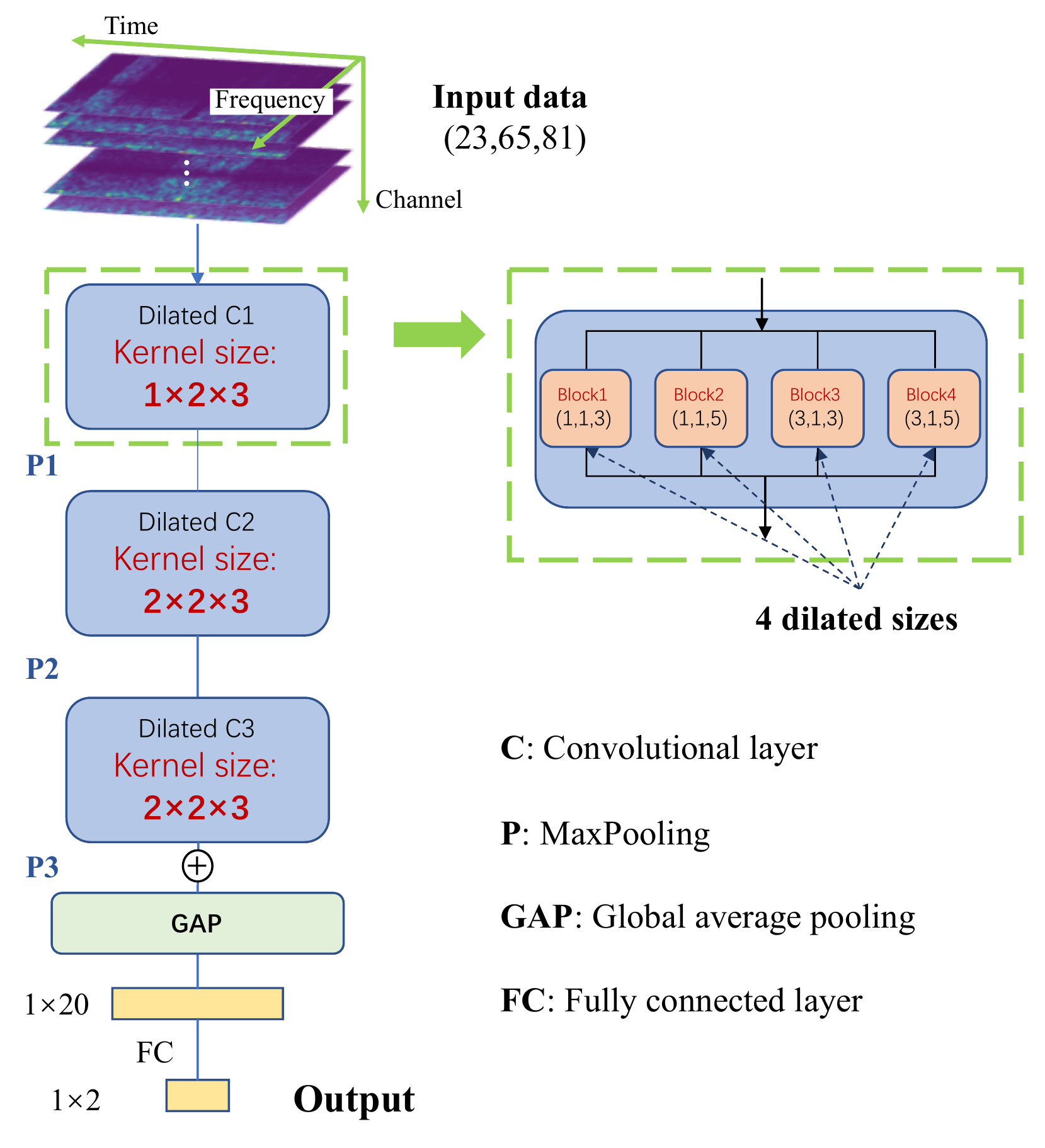}
		
		\caption{The structure of proposed dilated 3D CNN. C and P refer to convolutional and maxpooling layer respectively. Each convolutional layer consist of 4 blocks with different dilated size. The outputs of each block are combined together after the final layer, and then global average pooling (GAP) is used to reduce the amount of parameters. FC refers to a fully connected layer followed by a SoftMax function.}
		\label{model}
	\end{center}
\vspace{-0.3cm}	
\end{figure}

\begin{figure}[t]
	\begin{center}
		\subfigure[]{
			\includegraphics[width=4.0cm]{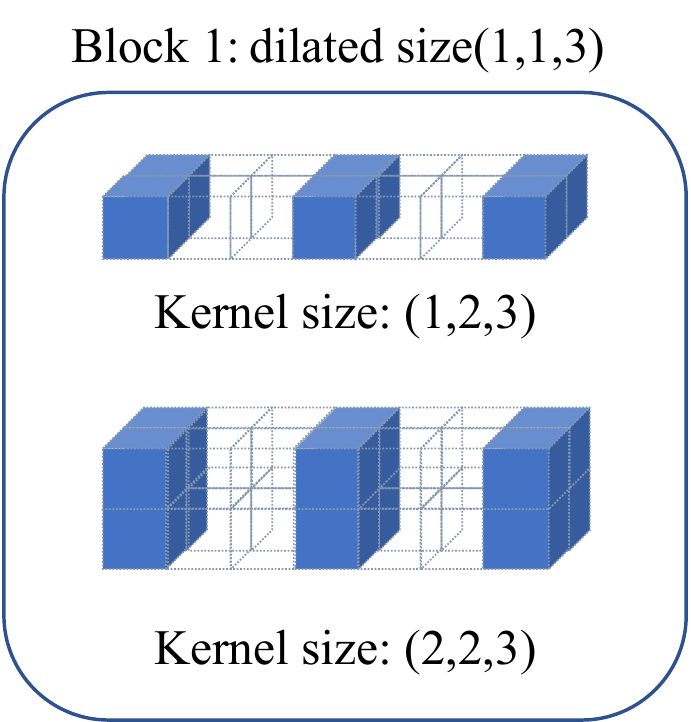}
		}
		\subfigure[]{
			\includegraphics[width=4.0cm]{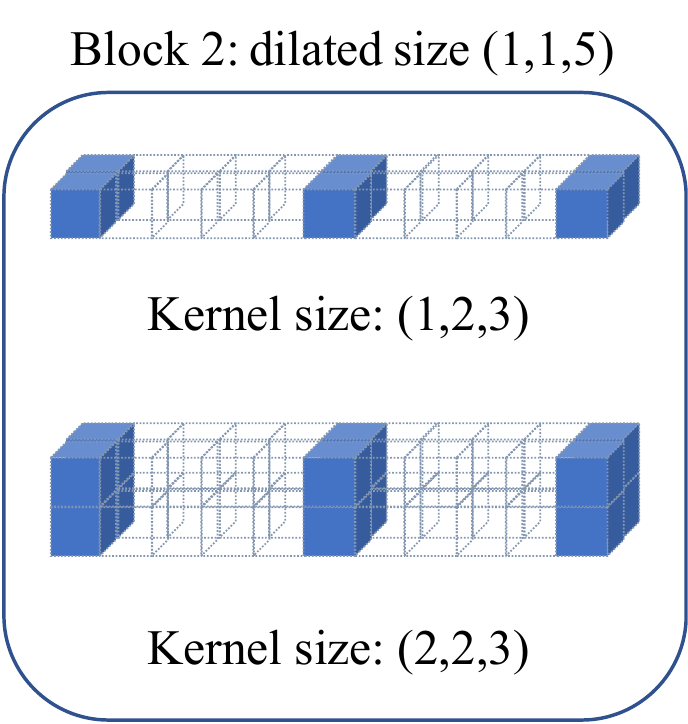}
		}
		\subfigure[]{
			\includegraphics[width=4.0cm]{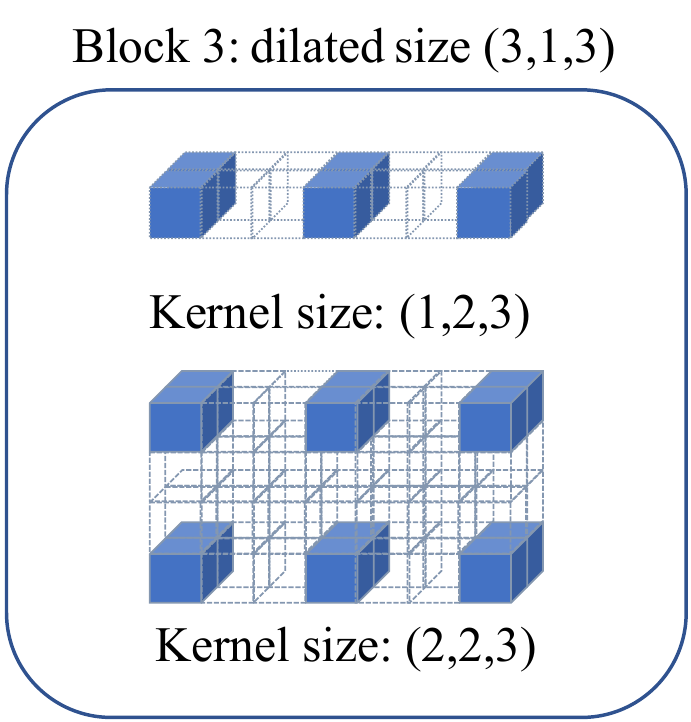}
		}
		\subfigure[]{
			\includegraphics[width=4.0cm]{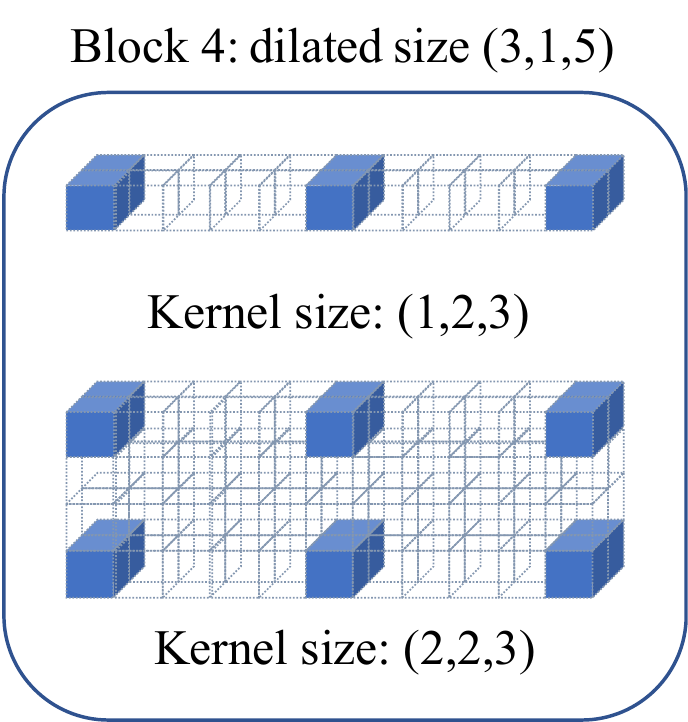}
		}
		\caption{The 3D convolution kernels with different dilated sizes: (a) (1,1,3), (b) (1,1,5), (c) (3,1,3), (d) (3,1,5).}
		\label{blocks}
	\end{center}
\end{figure}

The model structure is shown in Fig. \ref{model}. C and P refer to convolution and maxpooling layer respectively. The convolution kernel size of first layer is (1,2,3), the following layers are with the kernel size of (2,2,3). After each convolution layer there is a ReLU activate function and a maxpooling layer. The maxpooling kernel size is 1 $\times $ 2$\times $ 2 in P1, 2 $\times $ 2$\times $ 2 in P2 and P3. There are 4 blocks with different dilated size in the network, they are (1,1,3), (1,1,5), (3,1,3), (3,1,5) respectively (Fig. \ref{blocks}). The outputs of each block are combined together after the final layer. Global average pooling (GAP) is used to reduce the parameters, followed by a fully connected (FC) layer and a SoftMax function.

\begin{table}[t]
	\caption{The subjects information of CHB-MIT dataset.}
	\label{subjects}
	\small
	\centerline{
		\renewcommand\arraystretch{1.2}
		\begin{tabular}{p{1.0cm}<{\centering}p{2.5cm}<{\centering}p{3.5cm}<{\centering}}
			\hline
			\multirow{2}{*}{No.} & \multirow{2}{*}{Subject} & \multirow{2}{*}{\begin{tabular}[c]{@{}c@{}}Number of seizures/\\ Leading seizures\end{tabular}} \\
			&                          &                                                                                                 \\ \hline
			1                    & chb01                    & 7/3                                                                                             \\
			2                    & chb06                    & 10/6                                                                                            \\
			3                    & chb08                    & 5/3                                                                                             \\
			4                    & chb09                    & 4/3                                                                                             \\
			5                    & chb10                    & 7/6                                                                                             \\
			6                    & chb18                    & 6/3                                                                                             \\
			7                    & chb22                    & 3/3                                                                                             \\ \hline
		\end{tabular}
	}
	\vspace{-0.5cm}
\end{table}

\begin{table}[t]
	\caption{Comparison of sampling parameters, features and evaluation performance of main recently published models.}
	\label{comparison}
	\centering
	\renewcommand\arraystretch{1.2}
	\begin{threeparttable}
		\begin{tabular}[c]{p{2.3cm}<{\centering}p{1.0cm}<{\centering}p{1.1cm}<{\centering}p{0.6cm}<{\centering}p{0.6cm}<{\centering}p{0.6cm}<{\centering}}
			\hline
			Models                             & \begin{tabular}[c]{@{}c@{}c@{}}Window \\length/\\      Overlap\end{tabular}     & Features & {\scriptsize ACC} \tnote{1} & {\scriptsize TPR} \tnote{2} & {\scriptsize TNR} \tnote{3} \\ \hline
			Daoud \textit{et al.} \cite{RN19}           & 5 s/ 0                                                                      & Raw data & 0.719     & 0.718        & 0.720        \\
			Truong \textit{et al.} \cite{RN12}           & 30 s/ 8 s                                                                & STFT     & 0.753     & 0.815        & 0.691        \\
			Zhang \textit{et al.} \cite{RN18}             & 8 s/ 2 s                                                                       & PCC      & 0.727     & 0.756        & 0.698        \\
			Xu \textit{et al.} \cite{9073988}            & 20 s/ 5 s                                                                    & Raw data & 0.738     & 0.716        & {\scriptsize $\bm{ 0.760 }$}        \\
			Lawhern \textit{et al.} \cite{Lawhern_2018} & 5 s/ 0                                                                      & Raw data & 0.749     & 0.788        & 0.711        \\
			This work                          & 30 s/ 8 s                                                                  & STFT     & {\scriptsize $\bm{ 0.805 }$}    & {\scriptsize $\bm{ 0.858 }$}     & 0.751       \\ \hline
		\end{tabular}
		{\scriptsize \textsuperscript{1} Accuracy; \textsuperscript{2} True positive rate; \textsuperscript{3} True negative rate.}

	\end{threeparttable}
\vspace{-0.2cm}	
\end{table}
\section{Results}
\label{sec:result}
\subsection{Evaluation method and dataset}

To avoid overfitting and make this model more robust, in this work, leave-one-out cross validation method is used to simulate the situations in real life. Specifically, assuming that the EEG signals of a patient recorded N seizures in total, one seizure is selected as the testing set, the remaining N-1 are used to train the model, this process will be repeated N times. Furthermore, not all the seizures are available, we only focus on the prediction of leading seizures. Patients may have several seizures in a short period of time. Only the time interval between two seizures is greater than a certain value, they are considered as two different clusters. The first one of each cluster is called the leading seizure. Making distinction between leading seizure and the follow-up seizures is more meaningful for clinical research \cite{RN13}, but it also reduces the number of seizures, namely, the positive samples. After considering the trade-off and referring to previous studies, the time interval, namely, the seizure free time T is set as 4 h \cite{RN13} \cite{9073988}.

In this work, the model is evaluated with a public dataset called CHB-MIT \cite{doi:10.1161/01.CIR.101.23.e215}, which contains the scalp EEG recordings of 23 patients collected in Children’s Hospital Boston, with a sampling rate of 256Hz/s. Under the conditions that the leave-one-out cross validation method requires at least 3 seizures for each patient, we removed the subjects with missing data and selected 7 patients to do the experiments (Table \ref{subjects}).

\subsection{Comparison results}

We select main state-of-the-art contributions \cite{RN19} \cite{RN12} \cite{RN18} \cite{9073988} \cite{Lawhern_2018}, 
\begin{figure}[t]
	\begin{center}
		\subfigure[]{
			\includegraphics[width=7.8cm]{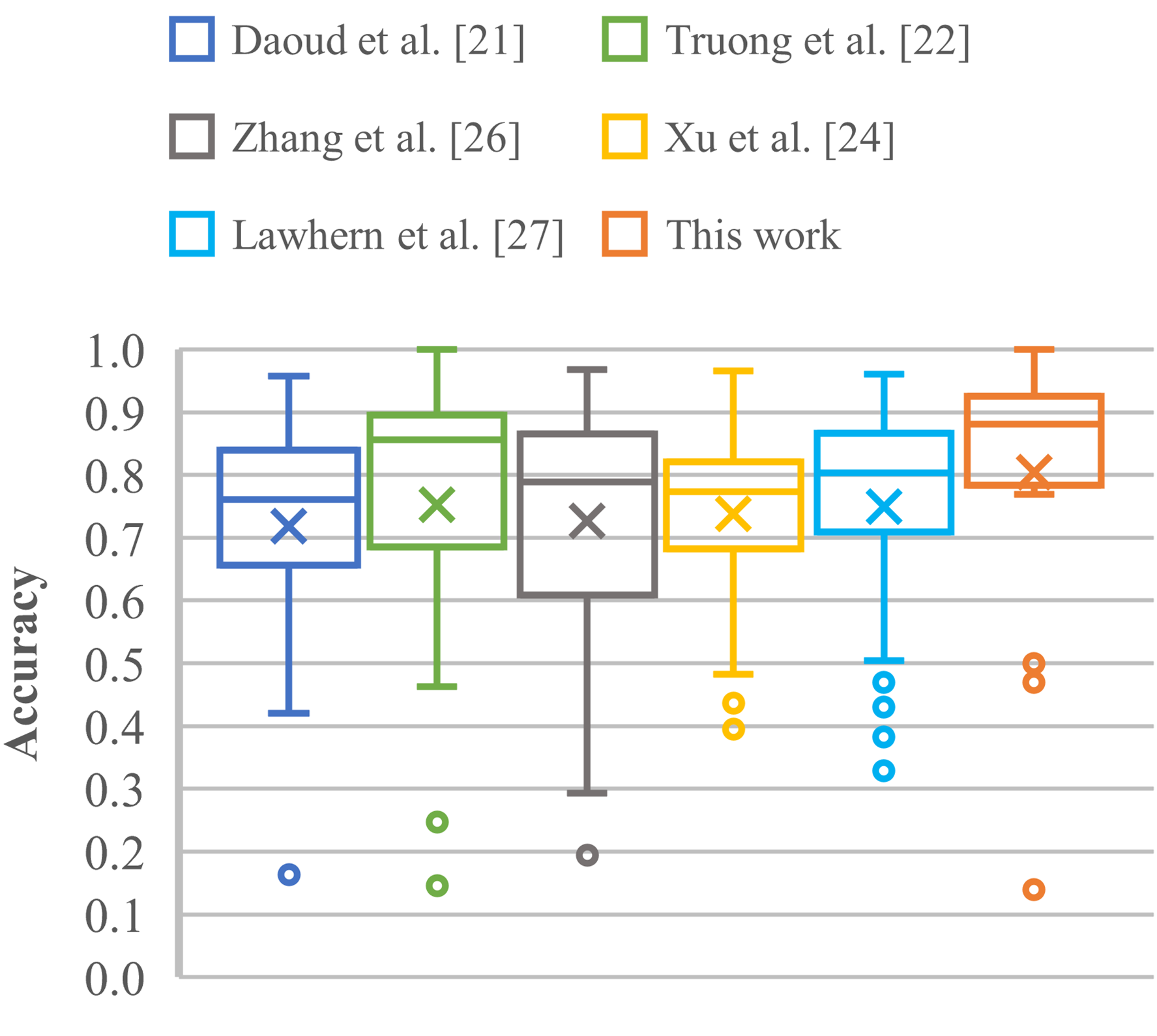}
		}
		\subfigure[]{
			\includegraphics[width=7.8cm]{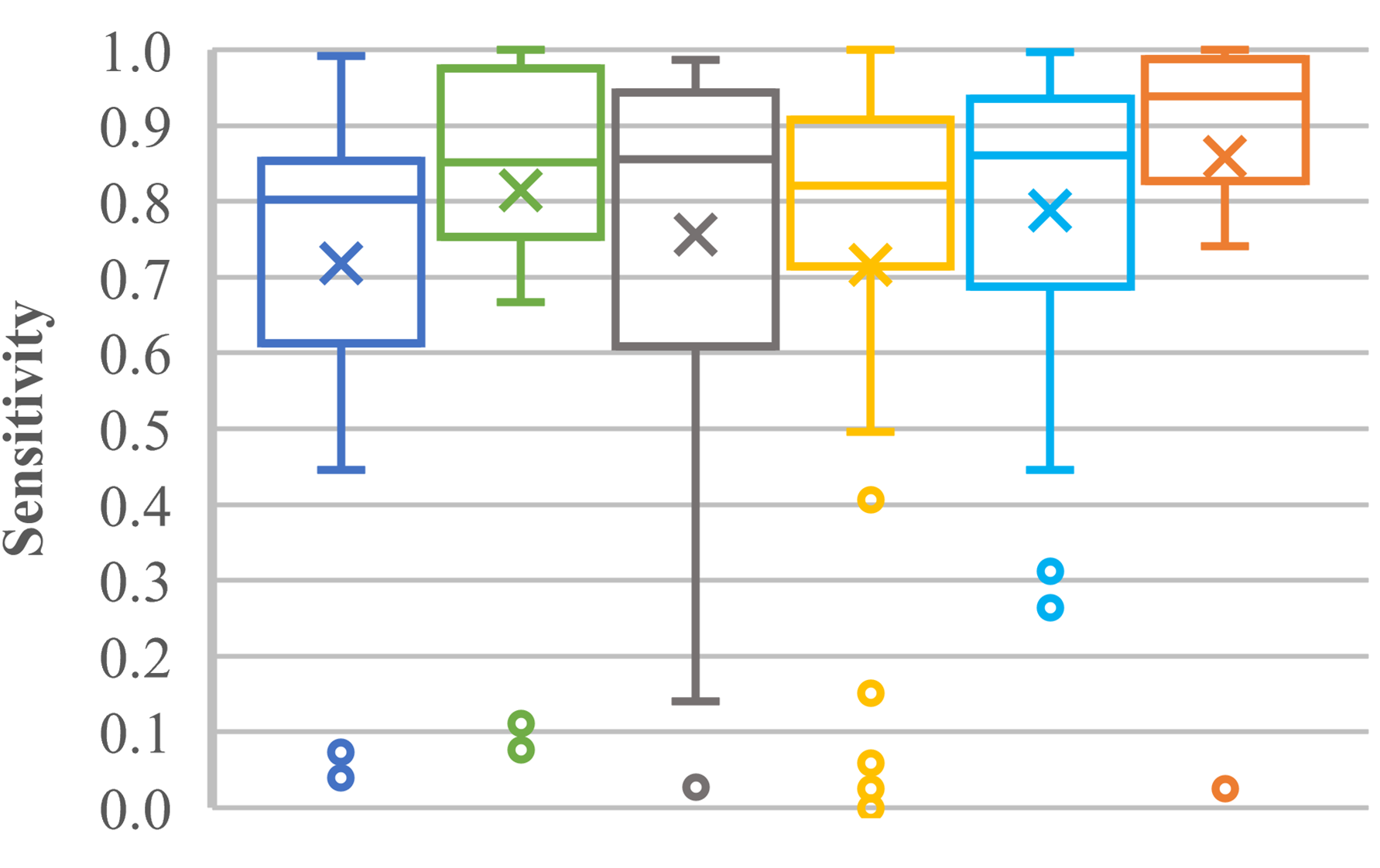}
		}
		\caption{ Performance comparison of different models: (a) Accuracy and (b) sensitivity.}
		\label{box_plot}
	\end{center}
	\vspace{-0.2cm}
\end{figure}
where the data preprocessing and model structure are described which allows us to reproduce their works for fair comparison. The data segmentation and dataset partition follows our proposed timings (SOP=30 min, SPH=5 min, T=4 h and leave-one-out cross validation adopted), sampling parameters and data preprocessing are consistent with the corresponding papers, which is shown in Table \ref{comparison}. Authors in \cite{RN19} \cite{9073988} and \cite{Lawhern_2018} used raw EEG data without any pre-processing as the models' input. Zhang et al. calculated Pearson correlation coefficient (PCC) to get the correlation coefficient matrix of each sample \cite{RN18}. Both contribution described in \cite{RN12} and our work used STFT to convert the raw EEG data into time-frequency domain. The evaluation results of each model are presented in Table \ref{comparison}, ACC, TPR and TNR refer to accuracy, true positive rate (i.e., sensitivity), true negative rate (i.e., specificity) respectively. Our model achieves the highest accuracy (80.5\%) and sensitivity (85.8\%), significantly higher than the 75.3\% accuracy and 81.5\% sensitivity of the second-ranked. Although 75.1\% specificity is not the highest, it is only slightly lower than 76.0\%, and the other two indicators of our model are much higher than \cite{9073988}. Figure \ref{box_plot} shows the overall accuracy and sensitivity of the cross validation experiments, our model has the smallest interval between the upper and lower edges of the box plot, which indicates that the performance of proposed model is more stable.

\section{Conclusion}
\label{sec:conclusion}

In this study, a novel multi-scale dilated CNN model was proposed for seizure prediction. STFT was applied to convert EEG signals into 3D tensors. Three-dimensional kernels were utilized to extract features from time, frequency and channel dimensions. Moreover, dilated convolution kernels enlarged the receptive field of the model and helped to obtain more abstract information from all dimensions. Evaluation results show that the proposed model achieves 80.5\% accuracy, 85.8\% sensitivity and 75.1\% specificity when performing seizure prediction task on the CHB-MIT database. Comparison results also indicate that the proposed model outperforms other state-of-the-art models.

\section*{Acknowledgment}

Authors would like to acknowledge funding support from the Westlake University and Zhejiang Key R\&D Program No. 2021C03002.

\bibliographystyle{IEEEbib}
\bibliography{my_references.bib}

\end{document}